\documentclass[letterpaper, 10 pt, journal, twoside]{IEEEtran}

\IEEEoverridecommandlockouts                              

\usepackage{graphics} 
\usepackage{graphicx} 
\usepackage{mathptmx} 
\usepackage{times} 
\usepackage{amsmath} 
\usepackage{amssymb}  
\usepackage{color}
\usepackage{subfig}
\usepackage{epstopdf}
\usepackage{wrapfig}
\usepackage{adjustbox}
\usepackage{booktabs, makecell}
\usepackage{multirow}
\usepackage{times}
\usepackage{textcomp}
\usepackage{floatrow}
\usepackage{float}
\usepackage[ruled, vlined]{algorithm2e}
\DeclareMathAlphabet{\mathcal}{OMS}{cmsy}{m}{n}
\usepackage[normalem]{ulem}
\usepackage[ruled]{algorithm2e}

\usepackage[dvipsnames]{xcolor}
\usepackage{hyperref}

\DeclareMathOperator*{\argmax}{arg\,max}
\DeclareMathOperator*{\argmin}{arg\,min}

\long\def\GeoFusion{{\em GeoFusion }}

\begin{document}

\title{
GeoFusion: Geometric Consistency informed \\
Scene Estimation in Dense Clutter
}

\author{Zhiqiang Sui, Haonan Chang,  Ning Xu, and Odest Chadwicke Jenkins%
\thanks{Manuscript received: March 1, 2020; Revised May 30, 2020; Accepted June 23, 2020.}
\thanks{This paper was recommended for publication by Editor Cesar Cadena upon evaluation of the Associate Editor and Reviewers' comments.
This work was supported by the Laboratory for Progress, University of Michigan, Ann Arbor} 
\thanks{Zhiqiang Sui, Haonan Changr,  Ning Xu, and Odest Chadwicke Jenkins are with the Department of Electrical Engineering and Computer Science, Robotics Institute, University of Michigan, Ann Arbor, MI, USA
        {\tt\footnotesize zsui@umich.edu}}%
\thanks{Digital Object Identifier (DOI): see top of this page.}
}

\let\oldtwocolumn\twocolumn
\renewcommand\twocolumn[1][]{%
    \oldtwocolumn[{#1}{
    \begin{center}
          \includegraphics[height =3.2cm]{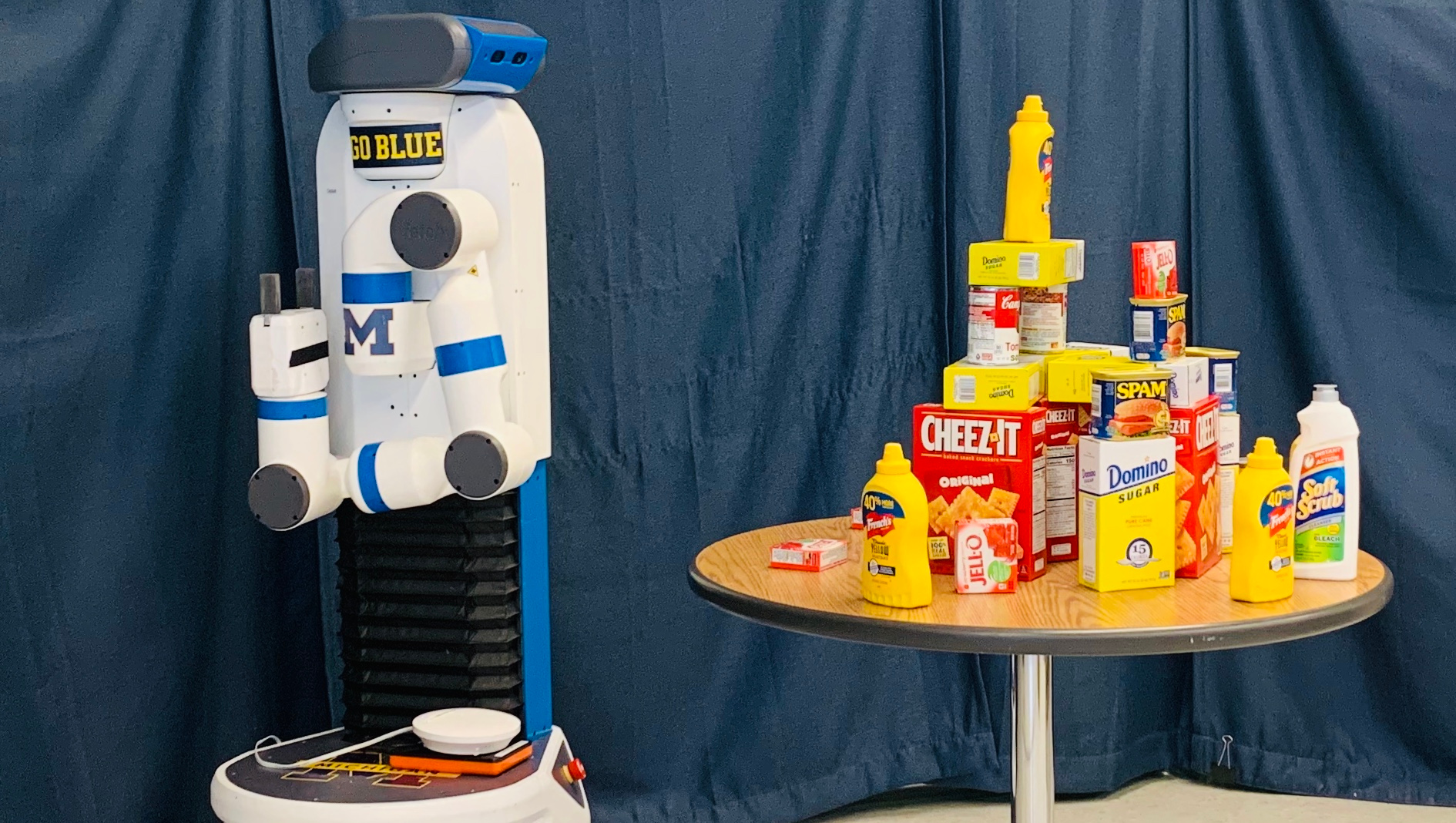} 
          \hspace{0.05cm}
          \includegraphics[height =3.2cm]{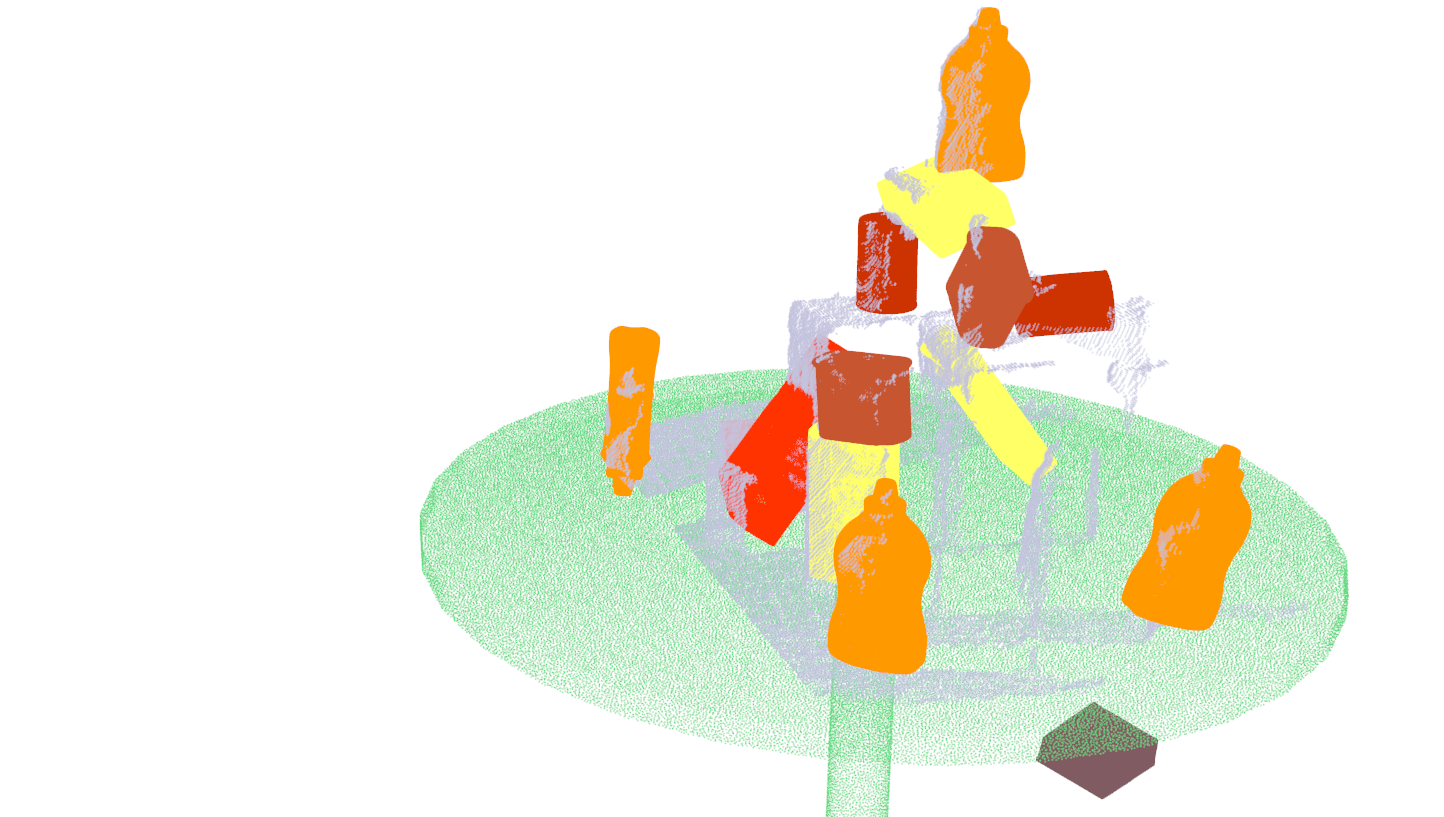}
          \hspace{0.05cm}
          \includegraphics[height = 3.2cm]{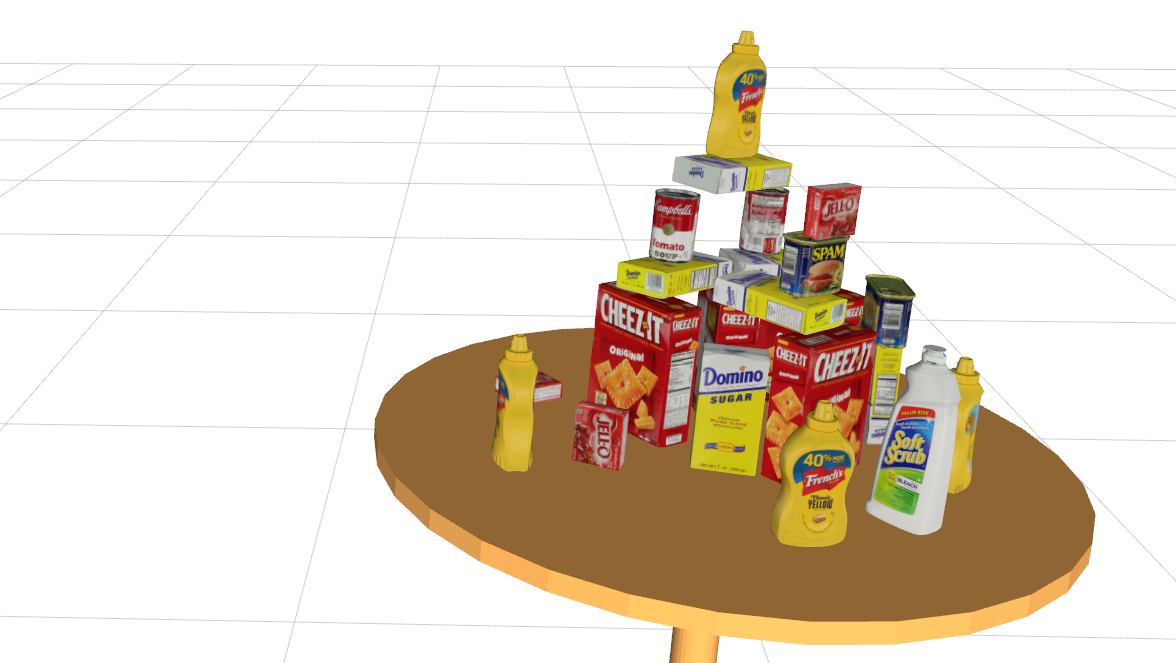}
          \captionof{figure}{\footnotesize In a densely cluttered tabletop scene (left) and given noisy semantic measurements (middle), the robot builds an object-level semantic map (right) .}
          \label{fig:teaser}
    \end{center}
    }]
}

\markboth{IEEE Robotics and Automation Letters. Preprint Version. July, 2020}
{Sui \MakeLowercase{\textit{et al.}}: GeoFusion: Geometric Consistency informed Scene Estimation} 

\maketitle

\begin{abstract}
We propose \textit{GeoFusion}, a SLAM-based scene estimation method for building an object-level semantic map in dense clutter.  In dense clutter, objects are often in close contact and severe occlusions, which brings more false detections and noisy pose estimates from existing perception methods. To solve these problems, our key insight is to consider geometric consistency at the object level within a general SLAM framework. The geometric consistency is defined in two parts: geometric consistency score and geometric relation. The geometric consistency score describes the compatibility between object geometry model and observation point cloud. Meanwhile, it provides a reliable measure to filter out false positives in data association. The geometric relation represents the relationship (e.g. contact) between geometric features (e.g. planes) among objects. The geometric relation makes the graph optimization for poses more robust and accurate. \GeoFusion  can robustly and efficiently infer the object labels, 6D object poses, and spatial relations from continuous noisy semantic measurements. We quantitatively evaluate our method using observations from a Fetch mobile manipulation robot.  Our results demonstrate greater robustness against false estimates than frame-by-frame pose estimation from the state-of-the-art convolutional neural network. 
\end{abstract}





\IEEEpeerreviewmaketitle

\section{introduction}

\IEEEPARstart{T}{o} make autonomous robots taskable such that they function properly, meet human expectations, and interact fluently with human partners, they must be able to perceive and understand the semantics of their environments~\cite{laird2017interactive}. More specifically, as robots move around in the environment, they must know what objects are presented as well as their locations. It is desired for robots to understand the semantic aspects of the scene and build a map at the object-level. 
An object-oriented representation of the world is a natural and efficient way for robots to make high-level decisions using task-level planners and help them to communicate with human users. 



The challenge is that many semantic aspects of the world are difficult for robots to sense directly due to the limited field-of-view and noisy observations from their onboard sensors.  Great progress has been achieved by the advances in deep neural networks for object detection \cite{ren2015faster, liu2016ssd, he2017mask} and 6D object pose estimation \cite{xiang2017posecnn, tremblay2018deep, wang2019densefusion}. 
However, building a semantic map at the object-level remains a challenging problem, especially in densely cluttered environments such as the one shown in Fig. \ref{fig:teaser}. Because in such dense clutter, objects are often in close contact, causing severe or complete occlusions.



One promising approach for building object-level maps is to fuse semantic measurements from different viewpoints. Robots can take advantage of their ability to move around the environment and provide continuous observations. 
More specifically, the data association between measurements and objects are first determined~\cite{wong2015data} and probabilistic model are built over objects and robot poses to infer the semantic map (SLAM++ \cite{salas2013slam++}, Fusion++ \cite{mccormac2018fusion++}) or belief over objects (CT-MAP \cite{zeng2018semantic}). Such methods perform inference directly on the object instance instead of low-level primitives (points, surfels).  This approach to inference offers faster inference, more compact map representation, and the potential for dynamic object reasoning. We posit that 
this approach is particularly advantageous 
in dense clutter. The geometric properties of object instances and geometric relationships between objects can be used for more robust data association and accurate pose estimation.



In this work, we present \textit{GeoFusion}, a SLAM-based approach for inferring object labels and 6D poses from continuous noisy measurements in dense clutter by exploring the \textit{geometric consistency} of the object instances. 
 The geometric consistency is defined in two parts: geometric consistency score and geometric relation. The geometric consistency score describes the compatibility between the object geometry model and the observation point cloud. Meanwhile, it provides a reliable measure to filter out false positives in data association. The geometric relation represents the relationship (e.g. contact) between geometric features (e.g. planes) among objects. Moreover, geometric relation is directly amenable to make better decisions for high-level task planners. 

Given noisy measurements from state-of-the-art object detection \cite{he2017mask} and pose estimation \cite{wang2019densefusion} systems, our method first determines the correct correspondences between measurements and objects with geometric consistency between measurements and point cloud observations. The associations are then used to build a factor graph along with the geometric relations between objects. We use the fast optimization technique to get the maximum likelihood estimation of object and robot poses. We quantitatively evaluate our method and demonstrate that \GeoFusion is able to estimate geometrically stable scene and be robust against false estimates from state-of-art frame-by-frame estimation system \cite{wang2019densefusion} and outperforms the baseline methods that do not consider geometric consistency.

\section{related work}


\subsection{Semantic Mapping}
From the emerging of semantic mapping \cite{kuipers2000spatial}, lots of work have explored this field with various semantic representations \cite{kostavelis2015semantic}. With the focus on the object-level semantic map, one widely used approach is to reconstruct a 3D map from either sparse features \cite{murAcceptedTRO2015} or dense point clouds \cite{newcombe2011kinectfusion, whelan2015elasticfusion, Rosinol20icra-Kimera} and then augments the map with the objects. Civera et al. \cite{civera2011towards} and Ekvall et al. \cite{ekvall2006integrating} used SURF/SIFT descriptors to register objects to the map created in a parallel thread. However, their work can not deal with objects in clutter. To deal with clutter, Li et al. \cite{li2016incremental} proposed an incremental segmentation approach to fuse segmentation across different frames based on dense reconstruction 3D map and used ObjRecRANSAC \cite{papazov2010efficient} to register object poses. But their method does not scale well with substantial false detections. 

Salas-Moreno et al. \cite{salas2013slam++} proposed an object SLAM system that recognized objects using Point-Pair Features in each frame and directly performed graph optimization on object and camera 6D poses. Their work shows promising results in the direction of building maps with 6D object poses. McCormac et al. \cite{mccormac2018fusion++} constructed an online object-level SLAM system with previously unknown shape and perform pose graph optimization at the object level. Both of their work assume independence between objects and hence difficult to work well in densely cluttered environments. Zeng et al. \cite{zeng2018semantic} proposed CT-MAP, which considers contextual relation between objects and temporal consistency of object poses in a conditional random field and maintain belief over object classes and poses. This approach is robust to false detections; however, does not scale well to the increasing number of objects due to the nature of pure generative inference. 




\subsection{Scene Estimation}
Great progress has been made by deep neural networks in object detection \cite{ren2015faster, he2017mask}, 6D object pose estimation \cite{xiang2017posecnn, wang2019densefusion} and semantic segmentation \cite{long2015fully}. To be robust to false detections due to the effects of overfitting, discriminative-generative algorithms \cite{narayanan2016discriminatively, sui2017sum} have been proposed for robust perception, especially in adversarial environments \cite{chen2019grip}. These efforts combine inference by deep neural networks  with sampling and probabilistic inference models to achieve robust and adaptive perception. However, the robustness is at the cost of computation time and the assumption object independence. 

Sui et al. \cite{sui2017goal} propose an axiomatic scene estimation method to estimate both geometric spatial relationships between objects and their 3D object poses for goal-directed manipulation. The spatial relation introduces strong constraints to reduce the search space of object poses. Desingh et al. \cite{desingh2016physically} and Mitash et al. \cite{mitash2019physics} leverage physics into scene estimation to search object poses based on compatibility score from physics simulation. The above-mentioned methods take either geometric and physical constraints into account for searching object poses, at the cost of computational efficiency.

\section{Problem Formulation}
Our aim is to estimate the semantic map composed of a collection $\mathcal{O} = \{ o_j \}_{j=1}^M$ of $M$ static objects as well as geometric relationships $\mathcal{R} = \{r_{ij} | i, j \in [1, M] \}$ between them with the assumption of known 3D object geometries. Note that the number of objects $M$ is unknown. Each object $o_j = (o_j^c, o_j^p)$ contains object class $o_j^c \in [1, C]$ and 6D object pose $o_j^p \in SE(3)$. Each $r_{ij}$ describes the geometric spatial relation between geometric features of two objects (e.g. support), more of which will be discussed at Sec. \ref{sec:geometric_relations}.

When the robot moves around in the environment, it observes a set of semantic measurements 
 $\mathcal{Z} = \{ \{z_t^k\}_{k=1}^{N_t}\}_{t=1}^T$, 
where $T$ is the total time step robot has travelled, 
and $N_t$  is the number of semantic measurements at each time step. Similar to the definition of object, each semantic measurement $z_t^k = (z_t^{k,c}, z_t^{k,p})$ is comprised of object class and 6D pose. The robot poses represent as $\mathcal{X} = \{x_t\}_{t=1}^T$, where $x_t \in SE(3)$ is also 6D pose in our case. In addition, the correspondence between objects $\mathcal{O}$ and measurements $\mathcal{Z}$ also needs to be determined and is defined as $\mathcal{D} = \{ \{d_t^k\}_{k=1}^{N_t} \}_{t=1}^T $ where $d_t^k = j$ stipulates that measurement $z_t^k$ corresponds to object $o_j$.

A complete statement of this problem is the maximum likelihood estimation of $\mathcal{X}$, $\mathcal{O}$, and $\mathcal{D}$ given the semantic measurements $\mathcal{Z}$ and the geometric relation $\mathcal{R}$ is computed heuristically from $\mathcal{O}$:

\begin{equation} \label{eq:overall-optimization}
    {\hat{\mathcal{X}}}, {\hat{\mathcal{O}}},  {\hat{\mathcal{D}}} = \argmax_{\mathcal{X}, \mathcal{O},  \mathcal{D}} \log p(\mathcal{Z} | \mathcal{X}, \mathcal{O},  \mathcal{D})
\end{equation}

As the joint estimation of objects, data association and robot poses suffers from the high dimensionality, 
the most common approach is to decompose Eq. \ref{eq:overall-optimization} into two separate estimation problems: data association and graph optimization.
The maximum likely estimation data association $\mathcal{D}$ is first computed given initial estimates of robot poses $\mathcal{X}^{(0)}$ and objects $\mathcal{O}^{(0)}$. Then given computed $\hat{\mathcal{D}}$, the most likely robot poses and objects are estimated:

\begin{align} 
    \hat{\mathcal{D}} &= \argmax_{\mathcal{D}}  p(\mathcal{D} | \mathcal{X}^{(0)}, \mathcal{O}^{(0)}, \mathcal{Z})  \label{eq:da} \\
       {\hat{\mathcal{X}}}, {\hat{\mathcal{O}}}  &= \argmax_{\mathcal{X}, \mathcal{O}} \log p(\mathcal{Z} | \mathcal{X}, \mathcal{O},  \hat{\mathcal{D}}) \label{eq:optimization}
\end{align}

Again, we decompose Eq. \ref{eq:optimization} to a two-step optimization where the object and robot poses are first optimized and the geometric relation $R$ will be computed heuristically from $O$ and these geometric relations will, in turn, pose constraints on factor graph for estimating geometrically stable scene: 
\begin{align} 
       {\hat{\mathcal{X}}}, {\hat{\mathcal{O}}}^{'} &= \argmax_{\mathcal{X}, \mathcal{O}} \log p(\mathcal{Z} | \mathcal{X}, \mathcal{O},  \hat{\mathcal{D}}) \label{eq:optimization-stage1} \\
       {\hat{\mathcal{O}}} &= \argmax_{\mathcal{O}} \log p(\mathcal{Z} | \mathcal{O}, \mathcal{R}) \label{eq:optimization-stage2} 
\end{align} 

\section{Data Association} \label{sec:da}

\begin{algorithm}[t]
\SetAlgoNoLine
\caption{Clustering Based Data Association}%
\label{alg:association}
\LinesNumbered 
\KwIn{Measurements $\mathcal{Z}$, Robot poses $\mathcal{X}$, Objects $\mathcal{O}$}
\KwOut{Data Associations $\mathcal{D}$}
\For {\textup{Each measurement $z_t^k$}}{
    \For{\textup{Each object $o_j$}}{
    \textup{Compute likelihood of $z_t^k$ of being object $o_j$:}\\
    $p(d_t^k=j) \propto  \mathbb I_{z^{k,c}_t =o_j^c} f_{gc}(z_t^k|o_j) f(z^{k, p}_t|o^{p}_j, x_t)$;\\
    }
    
    \eIf{$\max p_j \leq \epsilon_{new}$}
    {
        \textup{NewObjectInit($z_t^k$);}
    }
    {
        $d_t^k =\argmax p_j$;
    }
}

\For{\textup{Each object $o_j$}}{
    \textup{Compute false positive score} $f_j$ and remove it if score is above $\epsilon_{fp}$}

\end{algorithm}




As the number of objects $M$ is not known as the prior knowledge, we proposed a non-parametric clustering based approach for estimating data association in Eq. \ref{eq:da} based on DPmeans algorithm \cite{kulis2011revisiting}, as shown in Algorithm \ref{alg:association}.  Given initial estimates of objects $\mathcal{O}^{(0)}$, robot poses $\mathcal{X}^{(0)}$ and measurements $\mathcal{Z}$, we compute how likely each measurement belongs to current objects, being a new object or a false positive. The objects are created, updated, and deleted dynamically as the data association process moves forward.

\subsubsection{Association}


For each measurement $z^k_t$ in frame $t$, the likelihood of $z_t^k$ being assigned to object $o_j$ will be calculated as follows: 
\begin{equation}
    P(d_t^k=j) \propto  \mathbb I_{z^{k,c}_t =o_j^c} \, f_{gc}(z_t^k|o_j) \, f(z^k_t|o_j, x_t),
\end{equation}
where $\mathbb I_{z^{k,c}_t =c_i^c}$ is the label factor and it is obtained using an indicator function that evaluates to 1 if $z^{k,c}_t =c_i^c$ and 0 otherwise, meaning that each measurement will only be assigned to an object with the same label. 

\begin{figure}[t]
    \centering
   \includegraphics[width=0.9\textwidth]{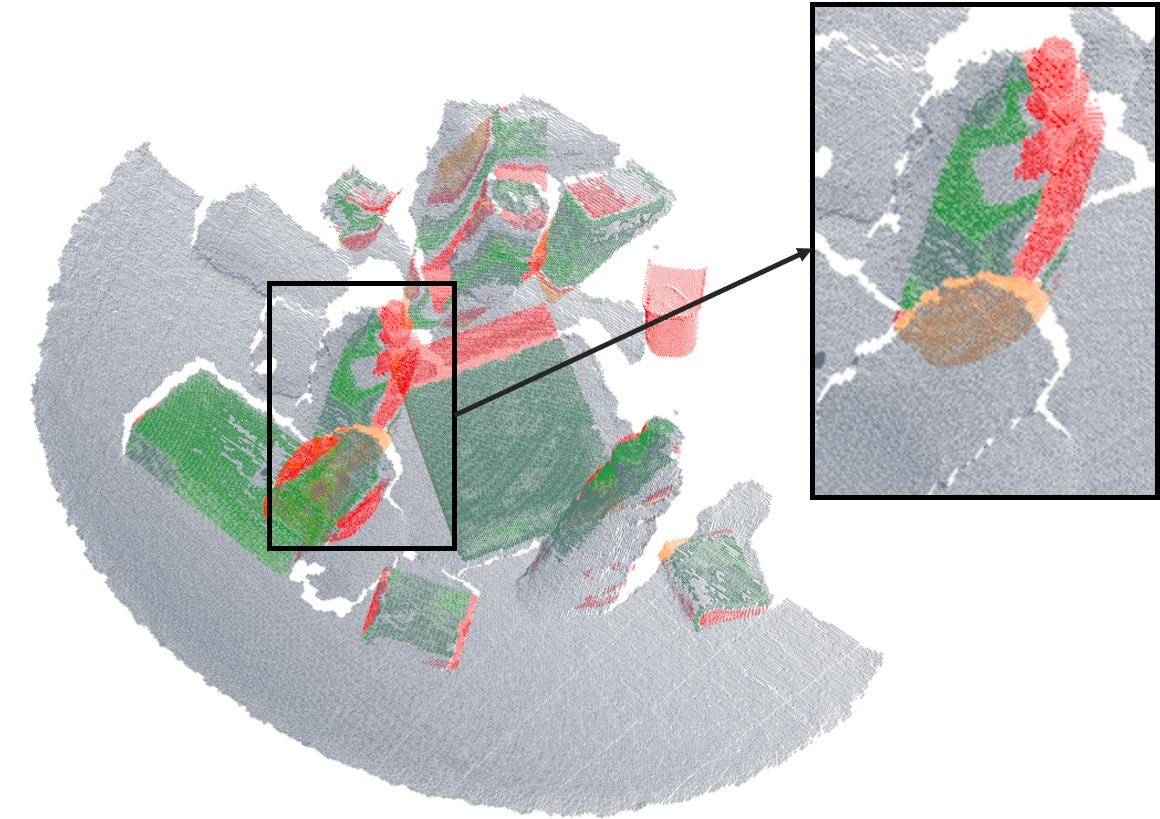}
    \caption{\footnotesize Geometric consistency score between measurements and point cloud observation. The zoomed picture shows the inlier, outlier and occlusion points of a mustard bottle. Point cloud observations are shown in dark blue. Inlier points are shown in green, outlier in red and occlusion in orange.} 
    \label{fig:inlier}
\end{figure}

Second term $f_{gc}(z_t^k | o_j)$ captures the geometric consistency score between measurement and the observation point cloud and provides a measure of how reliable the given measurement is. The motivation behind this term is that the confidence score given by neural networks sometimes are not accurate enough to describe the reliability of the measurement. Given a measurement $z_t^k$, a point cloud is back-projected from the rendered depth image and compared with observation point cloud to compute projective inlier ratio $r_i$, outlier ratio $r_{out}$ and occlusion ratio $r_{occ}$. A point in the rendered point cloud is
first considered as an occlusion point if it is occluded by the projective point in the observation point cloud shooting from the camera ray. If not, it is considered as an inlier if the distance with the projective point is less than certain sensor resolution $\epsilon_{res}$, or outlier if the distance is greater than $\epsilon_{out}$. Fig. \ref{fig:inlier} shows an example of different types of points. The ratios are computed over the total number of rendered points. And they are used as follows:

\begin{equation}
    f_{gc}(z_t^k | o_j) = S(r_{i}) S(1 - r_{out})S(1 - r_{occ})
\end{equation}
where $S$ is a modified sigmoid function providing an S-shaped logistic curve that is well-adapted to reflect the changing tendency of confidence on the given measurement over different values of each ratio.

The last term $f(z^k_t|o_j, x_t)$ is the pose factor that reflects the similarity of measurement pose and object pose, given the current robot pose $x_t$. Here, we make the assumption that the likelihood of each measurement $z_t^k$ given the robot pose and the object pose follows a multivariate Gaussian distribution.
\begin{equation}
    f(z_t^k|o_j, x_t) \sim \mathcal{N}(x_t^{-1}\cdot o_j^p , Q)
\end{equation}
where $Q$ is the measurement noise matrix and this factor can thus be drawn from the Gaussian probability density function. 
With all the factors described above, $z_t^k$ is assigned to be the maximum likelihood object if the object is within a certain threshold $\epsilon_{new}$, otherwise, it is assigned to a new object.


The One Measurement Per Object (OMPO) constraint \cite{wong2015data} is also applied in this process so that two measurements in the same frame will never be assigned to the same object and the latter will be assigned to a new object instead.

\subsubsection{False positive removal}
For each object candidate created or updated in the association step, we compute its false positive score $f_j$ as follows: 
\begin{equation}
f_j = 1 - \frac{R_j}{1+e^{(-n_j)}}  
\end{equation}
where $n_j$ is the number of measurements assigned to $o_j$ and $R_j$ is the maximum geometric consistency score among all of these measurements. 
If there are more measurements assigned to one object, it is reasonable to consider this object as more reliable among others thus it deserves a lower false-positive score. All objects with false-positive scores higher than $\epsilon_{fp}$ will then be deleted.


\begin{paragraph}{Overlapped objects merging}  
 We notice that most of the false positive measurements usually overlap with other objects, especially for symmetric objects, like a cylinder-shape tomato soup can. Object-level geometric consistency is incorporated again to address this problem as we applied Separating Axis Theorem (SAT) 
 to detect the collision ratio of the three-dimensional bounding boxes of any two objects. Corner points of each bounding box are projected to several main axes and the overall collision ratio is decided based on the percentage of overlap along each axis. One object with a high collision ratio will be merged to the other one that has a lower false-positive score.

 \end{paragraph}

\section{Graph Optimization} \label{sec:optimization}
\begin{figure}[t]
    \centering
    \includegraphics[width=1.0\linewidth]{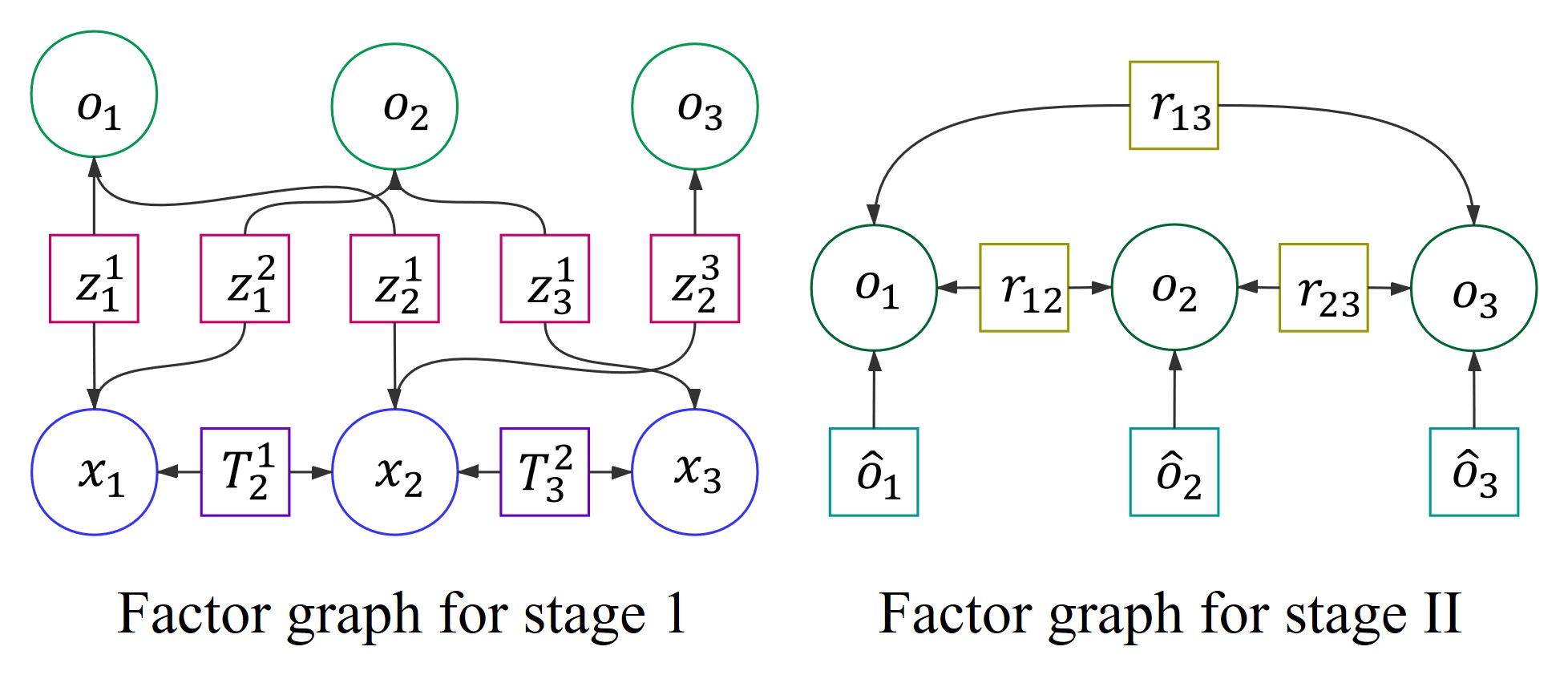}
    \caption{\footnotesize Factor graphs for our two stage optimization method.}
    \label{factor graph}
\end{figure}


After the associations $\mathcal{D}$ between objects $\mathcal{O}$ and semantic measurements $\mathcal{Z}$ have been determined, the most likely robot and object poses are then inferred in the graph optimization process. The factor graph is used to model robot and object states and expresses the conditional independence between them. It gained a huge success in solving SLAM problems \cite{dellaert2017factor} due existing computational tools that allow efficient optimization \cite{dellaert2012factor}. In a factor graph, there exists a set of vertices $\mathcal{V}$ that represent optimization random variables and edges represent factors $\mathcal{F}$ among a subset of random variables. A factor $f$ represents probabilistic dependencies among the random variables and is associated with a cost function. Graphically, the vertex is a circle and the factor is a square in the factor graph, as shown in Fig. \ref{factor graph}. The joint probability of the optimization random variables can be expressed as the product of factors: 

\begin{equation}
    p(\mathcal{V}) \propto \prod_{f \in \mathcal{F}} f(\mathcal{V})
\end{equation}

The optimization process in \GeoFusion is decomposed into two stages, as shown in Fig. \ref{factor graph}. First, 6D pose of the robot and the objects are optimized over odometry and semantic measurements. In the second stage, 6D pose of objects are then fine-tuned over the computed geometric relations to produce a geometric consistent scene estimate.

\subsection{Stage I Optimization}
The right hand side of Eq. \ref{eq:optimization-stage1} is rewritten as:

\begin{equation}
     \log p(\mathcal{Z} | \mathcal{X}, \mathcal{O},  \hat{\mathcal{D}}) 
     = \sum_{t = 1}^T \phi (T_t^{t-1} ; x_{t - 1}, x_t)  + \sum_{t = 1}^T \sum_{k = 1}^{N^t} \phi (z_t^{k, p} ; x_t, o_{d_t^k}^p) \label{eq:optimization-stage1-log} \nonumber
\end{equation}
where $\phi (T_t^{t-1} ; x_{t - 1}, x_t)$ is the odometry factor and $\phi (z_t^{k, p} ; x_t, o_{d_t^k}^p)$ is the object measurement factor. $T_t^{t-1}$ is the odometry measurement between robot pose $x_{t-1}$ and $x_{t}$. $z_t^{k,p}$ is the semantic measurement of an object between the robot pose $x_t$ and the object pose $o_{d_t^k}^p$. With the standard assumption of additive Gaussian noise, each factor follows a quadratic form:

\begin{align}
    \phi (T_t^{t-1} ; x_{t - 1}, x_t) &= - \frac{1}{2} (x_t \ominus x_{t-1} - T_t^{t-1}) Q^{-1} (x_t \ominus x_{t-1} - T_t^{t-1}) \nonumber \\
    \phi (z_t^{k, p} ; x_t, o_{d_t^k}^p) &= - \frac{1}{2} (x_t \ominus o_{d_t^k}^p - z_t^{k, p}) R^{-1} (x_t \ominus o_{d_t^k}^p - z_t^{k, p}) \nonumber
\end{align}
where $ \ominus $ represents the operator that computes the relative transformation. $Q$ and $R$ is the corresponding covariance noise matrix. Thus, The maximum likelihood estimation of $X$ and $O$ can be written as the nonlinear least-squares problem:

\begin{align}
    {\hat{\mathcal{X}}}, {\hat{\mathcal{O}}}^{'} &= 
     \argmin_{\mathcal{X}, \mathcal{O}} \sum_{t=1}^T \| x_{t-1} \ominus x_t - T_{t}^{t-1} \|_{\Sigma_Q} \nonumber \\
     &\quad + \sum_{t=1}^T \sum_{k-1}^{N_t} \| x_t \ominus o_{d_t^k}^p - z_t^{k, p} \|_{\Sigma_R} \footnotemark \label{eq:stage1-least-square} 
\end{align} \footnotetext{$\| X \|_{\Omega} = \sqrt{X^T \Omega X} $}
where $\Sigma_Q$ and $\Sigma_R$ is the information matrix. Note that the as the metrics for translations and rotations are different, the information matrix for them is also different. We use $\Sigma_{p}$ for translations and $\Sigma_{q}$ for rotations which represent as quaternions. $\Sigma_p$ and $\Sigma_{q}$ are diagonal matrices and represent as $Diag(\omega_p, \omega_p, \omega_p)$ and $Diag(\omega_q, 0, 0, 0)$. For symmetric objects in the object measurement factor, we design a separate $Sigma_{q}$ for the rotation. For example, object with rotation axis in z-axis is associated with $\Sigma_q = Diag(0, \omega_q,\omega_q, 0) $ to exclude the influence from z-axis.

\begin{figure}
    \centering
    \includegraphics[width=0.8\linewidth] {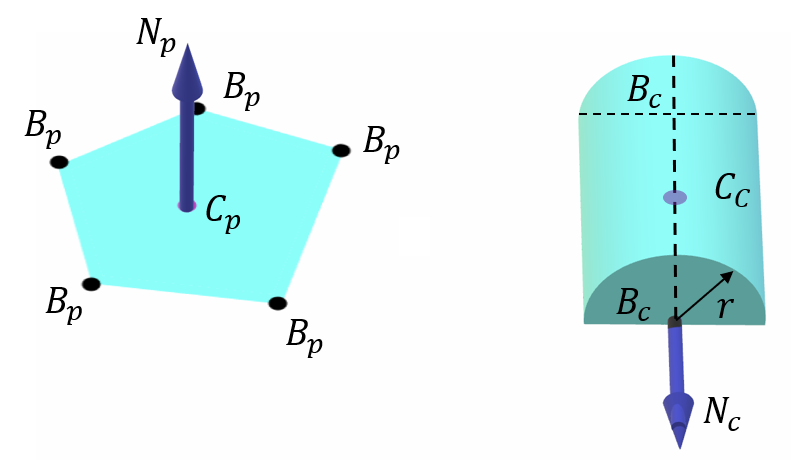}
    \caption{\footnotesize Two types of surface feature: the left one is plane surface feature, and the right one is curved surface feature.}
    \label{surface feature}
\end{figure}
\subsection{Object Geometric Relationship Inference} \label{sec:geometric_relations}
In stage I optimization, the maximum likelihood estimation of robot and object poses is obtained by minimizing the odometry and object measurement cost. However, the resulting object poses are still noisy and not accurate enough for applications like precise robot manipulation. The noisy object poses lead to geometrically inconsistent scene estimate, e.g., floating or intersection between objects. To get a highly accurate and geometrically consistent scene estimate, we posit that if the geometric relationship between objects can be inferred and these relations can serve as the constraints (factors) between objects in the factor graph model, as shown in the right of Fig. \ref{factor graph}. Given estimated object poses $\mathcal{O}$ from Eq. \ref{eq:stage1-least-square}, $\mathcal{R}$ is heuristically computed.  

\subsubsection{Geometric Surface Feature} 
For common rigid household objects, we only consider \textit{contact} relations between surfaces for generality in representing the geometry of the object and efficiency in the graph optimization process. In this paper, geometric surface features are divided into two classes: plane surface feature and curved surface feature. An illustration of surface features is shown in Fig \ref{surface feature}.

Given an arbitrary convex polyhedron, geometric surface features can be extracted. For example, there are six plane surfaces and twelve curved surfaces for a cube-like geometry as it has six planes and twelve edges. The edge is treated as a zero radius curve surface. For cylinder-like geometry, there are two plane surfaces and one curved surface. Although the two classes of surfaces cannot represent all of the objects (e.g., articulated or non-rigid objects), they are general and sufficient for common rigid household objects.






\begin{paragraph}{Plane Surface Feature}
The plane surface feature refers to those planes on polyhedrons. There are three attributes associated with plane surface feature: a center point $C_p \in R^3$, a set of points $B_p = \{b^i_p\}_{i=1}^{N_b}$ with $b^i_p \in R^3$, and the normal direction $N_p \in R^3$. $C_p$ describes the 3D position of the plane center. Each $b^i_p$ is the 3D position of a boundary point on the surface and $N_b$ is the number of boundary points.  

\end{paragraph}

\begin{paragraph}{Curved Surface Feature}
The curved surface feature describes curved surfaces in 3D shapes like cylinders. Noticeably, edges of polyhedrons are also regarded as a curved surface feature with a radius equal to zero. Five attributes are associated with the curved surface feature: $C_c \in R^3$ and $N_c \in R^3$, determining the 3D position of center and direction of the rotation axis of the curved surface. $B_c = \{b_c^i\}_{i=1}^2$ with $b_c^i \in R^3$ refers to the 3D position of boundary points in rotation axis. Radius $r$ describes the radius of the curved surface.
\end{paragraph}

\begin{figure*}[t]
    \centering
    \includegraphics[width=0.8\linewidth] {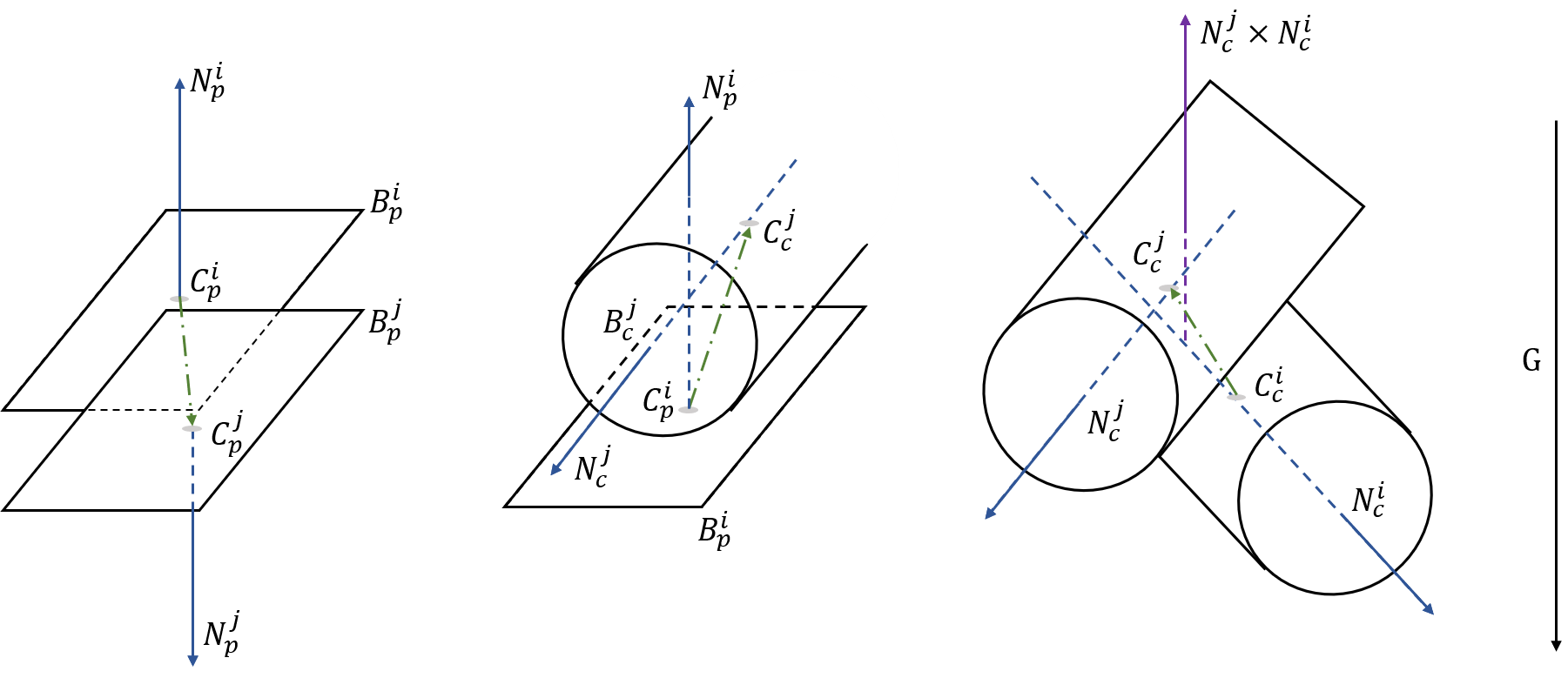}
    \caption{\footnotesize Illustrations for geometric relationships. Plane to plane (P2P) contact, plane to curved surface (P2C) contact, and curved to curved surface (C2C) contact are shown from the left to the right. }
    \label{geometric relationship}
    \vspace{-2mm}
\end{figure*}

Accordingly, we define three types of spatial relations between these two surfaces: Plane2Plane Contact (P2P), Plane2Curve Contact (P2C), and Curve2Curve Contact (C2C). Fig \ref{geometric relationship} illustrates these geometric relations.


\subsubsection{Geometric Feature Relationship Inference}
\begin{paragraph}{P2P contact}
If two different plane surface features $P_u,\ P_v$ from two objects $o_i,\ o_j$ are contacting each other, their attributes should follow the following rules:
\begin{equation}
\begin{array}{c}
|T_{o_i}N_{P_u}^i \cdot T_{o_j}N_{P_v}^j + 1| < \epsilon_n^{pp} \\
|T_{o_i}N_{P_u}^i \cdot (T_{o_j}C_{P_v}^j - T_{o_i}C_{P_u}^i)| < \epsilon_c^{pp} 
\end{array}
\end{equation}
where $\epsilon_n^{pp}, \epsilon_c^{pp}$ are the direction threshold and distance threshold for P2P contact. 
\end{paragraph}

\begin{paragraph}{P2C contact}
If a plane surface feature $P_u$ and a curved surface feature $C_v$ are contacting each other, their attributes should follow the following rules:
\begin{equation}
\begin{array}{c}
|T_{o_i}N_{P_u}^i \cdot  T_{o_j}N_{C_v}^j| < \epsilon_n^{pc} \\
|T_{o_i}N_{P_u}^i \cdot (T_{o_j}C_{C_v}^j - T_{o_i}C_{P_u}^i) - r_j| < \epsilon_c^{pc}
\end{array}
\end{equation}
where $\epsilon_n^{pc}, \epsilon_c^{pc}$ are the direction threshold and distance threshold for P2C contact.
\end{paragraph}

\begin{paragraph}{C2C contact}
If two different curved surface features $C_u,\ C_v$ are contacting each other, their attributes should follow the following rules:
\begin{equation}
   |(T_{o_i}N_{C_u}^i \times T_{o_j}N_{C_v}^j) \cdot (T_{o_j}C_{C_v}^j - T_{o_i}C_{C_u}^i) - (r_i + r_j)| < \epsilon_c^{cc}
\end{equation}
where $\epsilon_c^{cc}$ is the distance threshold for C2C contact. 
\end{paragraph}


\subsubsection{Physical Relationship Check}
We also add two physical based check in complement to the above geometric methods to ensure eliminating false detected contact. For example, if two cubes are placed parallel and close to each other on the table, a P2P contact can be detected. However, there is a chance that those objects are just near to each other but not contacting each other.



\begin{paragraph}{Support direction check}
If a plane feature can support another plane or be supported, its normal direction can not be horizontal to the gravity direction. Assume the direction of gravity is $N_G$, the support direction check of plane feature $i$ is defined as:
\begin{equation}
    N_G \cdot N_p^i > \epsilon_G
\end{equation}
where $\epsilon_G$ is the threshold for support direction check.
\end{paragraph}
Here we make an implicit assumption that there is at least one grounded object, e.g. table, which supports all the other objects. Similar assumption can be found in SLAM++ \cite{salas2013slam++}.

\begin{paragraph}{Qualitative Support projection check}
If there is a support relationship between two features, their 2D projections along the direction of gravity must be overlapped. For those contact feature candidates, their 2D projections are the polygon or the edge projected from their boundary points. Separate axis theorem (SAT) is applied to check the overlapping of those 2D contours. Those feature candidates whose 2D projections are not overlapped will be removed.
\end{paragraph}

\subsection{Stage II Optimization}
After spatial relations $\mathcal{R}$ between objects have been computed from the estimated object poses in stage I optimization, these relations are used as constraints in the factor graph optimization. These constraints will enforce proper contact between objects and fix the issues of floating and interpenetration. We can rewrite the right hand side of Eq. \ref{eq:optimization-stage2} as:

\begin{equation}
     \log p(\mathcal{Z} | \mathcal{O}, \mathcal{R}) = \sum_{i=1}^M \sum_{j=1}^M \phi (r_{ij} ; o_i^p, o_j^p) \sum_{i=1}^M \phi(o_i^{'} ; o_i)
\end{equation}
where $\phi (r_{ij} ; o_i^p, o_j^p)$ is the geometric relation factor and $\phi(o_i^{'} ; o_i)$ is the object prior factor. The object prior factor comes from the estimated object poses in stage I optimization. The geometric relation factor encodes the contact constraint between two objects computed from the previous section. Accordingly, there are three constraints: P2P, P2C, and C2C. For P2P constraint, the cost function is defined as: 
\begin{equation}
    E_{P2P} =  \omega_q |T_{o_i}N_{P_u}^i \cdot T_{o_j}N_{P_v}^j + 1|  +  \omega_p
|T_{o_i}N_{P_u}^i \cdot (T_{o_j}C_{P_v}^j - T_{o_i}C_{P_u}^i)| \nonumber
\end{equation}
Similarly, the cost function of P2C constraint is defined as:
\begin{equation}
    E_{P2C} =  \omega_q | T_{o_i}N_u^{o_i} \cdot T_{o_j}N_v^{o_j} | 
    + \omega_p |T_{o_i}N_u^{o_i} \cdot (T_{o_i}C_u^{o_i} - T_{o_j}C_v^{o_j})| \nonumber
\end{equation}
and the cost function of C2C is defined as:
\begin{equation}
    E_{p2c} = \omega_p |(T_{o_i}N_u^{o_i} \times T_{o_j}N_v^{o_j}) \cdot (T_{o_i}C_u^{o_i} - T_{o_j}C_v^{o_j}) 
    - (r_v + r_u)| \nonumber
\end{equation}

\section{Experiments}

\begin{table}
    \centering
    \begin{center}
\begin{tabular}{ c|c|c|c|c } 
 \toprule
 & FbF & B-SLAM & R-Front & \GeoFusion  \\
 \midrule
$\text{mAP}_{50}$ & 58.4 & 56.0 & 72.3 & \textbf{73.4} \\ 
 \hline
$\text{mAP}_{75}$ & 28.6 & 24.6 & 50.3 & \textbf{59.5} \\ 
\hline
$\text{mAP}_{50:95}$ & 30.5 & 27.9 & 45.2 & \textbf{50.3} \\ 
 \hline
 
\end{tabular}
\end{center}
    \caption{\footnotesize Object detection performance of different methods with mAP under different IoU.  }
    \label{tab:object-detection}
\end{table}

\begin{table} 
    \footnotesize
    \centering
    \begin{center}
\begin{tabular}{c |cc|cc|cc|cc}
\toprule
  & \multicolumn{2}{c|}{FbF} & \multicolumn{2}{c|}{B-SLAM} & \multicolumn{2}{c|}{R-Front}  & \multicolumn{2}{c}{\GeoFusion} \\
   & Pr & Rec & Pr & Rec & Pr & Rec & Pr & Rec \\

\midrule
 $\text{mAP}_{50}$ & 0.87 & \textbf{0.60} & 0.87 & 0.27 & 0.91 & 0.43 & \textbf{0.99} & 0.51 \\
 \hline
 $\text{mAP}_{75}$ & 0.56 & 0.38 & 0.63 & 0.19 & 0.75 & 0.35 & \textbf{0.91} & \textbf{0.46} \\
 \hline
 $\text{mAP}_{50:95}$ & 0.52 & 0.35 & 0.57 & 0.17 & 0.65 & 0.30 & \textbf{0.77} & \textbf{0.40} \\ 
 \hline
\end{tabular}
\end{center}
    \caption{\footnotesize Object detection performance on precision (Pr) and recall (Rec) for different methods under different mAP. The detection confidence threshold is set to 0.5. }
    \label{tab:object-detection-precision-recall}
\end{table}

\subsection{Implementation}
We use Mask R-CNN \cite{he2017mask} and DenseFusion \cite{wang2019densefusion} to generate semantic measurements where the former detects objects along with the instance segmentation which the latter takes in to estimate the 6D pose. We finetune the Mask R-CNN with the YCB-Video Dataset \cite{xiang2017posecnn} with the pretrained Microsoft coco model \cite{lin2014microsoft}. We use the public available DenseFusion weights and implementation without fine-tuning. The front-end in our implementation selects every 10th camera frame as a keyframe. The camera visual odometry is provided by ORB-SLAM2 \cite{murAcceptedTRO2015}. We use Ceres \cite{ceres-solver} as the optimization backend. In our implementation, we perform stage I optimization every 10th keyframe and perform stage II optimization every keyframe.


\subsection{Dataset and Baseline}
To test the performance of our method, we collect a testing dataset \footnote{ \url{http://www-personal.umich.edu/~zsui/geofusion_dataset.html} } from six RGBD video streams in which our Michigan Progress Robot moves around the table. In each scene, we put around 18 objects in dense clutter on the table and collect around 200 keyframe RGB-D observations from the robot's sensor. 
The groundtruth object classes and poses are labelled using LabelFusion \cite{marion2018label}. For baseline methods, we compare our method with frame-by-frame (FbF) estimation from Mask R-CNN and DenseFusion. It only considers the single frame and does not take any temporal or spatial constraints into account.  We also compare with the variations of our proposed \textit{GeoFusion}: 1) a baseline SLAM (B-SLAM) method without considering geometric consistency in both data association and graph optimization. The confidence score from DenseFusion instead of the geometric consistent score is used in computing data association. The back-end graph optimization only optimizes over odometry and object poses without considering spatial constraints between objects.  2) a robust front-end (R-Front) method where only our data association is used and not the graph optimization.

\begin{figure}[t]
    \centering
        \includegraphics[width=0.9\columnwidth]{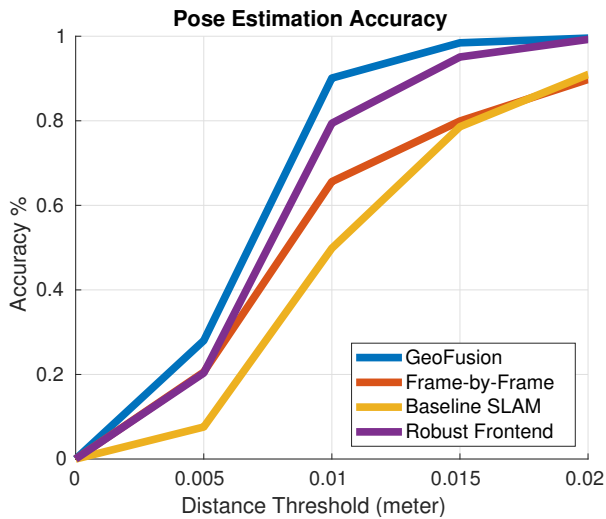}
       
    \caption{\footnotesize Pose accuracy comparison between different methods. }
    \label{fig:pose_acc}
    \vspace{-2mm}
\end{figure}

\subsection{Evaluation} 
\subsubsection{Object Detection}
We first compare object detection performance with baseline methods on each keyframe collected in the dataset. 

\begin{paragraph}{mAP}
We use the common object detection metric: mean average precision (mAP), to evaluate object detection. Average precision (AP) is the area under the Precision-Recall curve of an object class and mean average precision is the mean of all the AP. The subscript under mAP is the ratio of intersection over union (IoU) between the estimated bounding box and the ground truth bounding box. 
We also use mAP@[50:95] which corresponds to the average AP for IOU from 0.5 to 0.95 with a step size of 0.05.  

Results are shown in Table \ref{tab:object-detection}. \GeoFusion significantly outperforms the frame-by-frame method in all mAP metrics indicating that our method is capable of identifying false detections and keeping true ones from noisy semantic measurements. Robust front-end method achieves the similar $\text{mAP}_{50}$ with \GeoFusion however underperforms in $\text{mAP}_{75}$ and $\text{mAP}_{50:95}$. It shows that the back-end optimization improves the 6D poses of objects, which in turn increases the mAP with a tighter threshold in IOU. The baseline SLAM method does not take any geometric consistency into account and the performance is even worse than the frame-by-frame method. This is because the false and noisy measurements accumulate in the baseline slam and without the help of geometric consistency, they are difficult to be eliminated from the estimates. 
\end{paragraph}

\begin{paragraph}{Precision and Recall}
For a more intuitive evaluation on object detection, we also report numbers on precision (Pr) and recall (Rec) with the confidence score threshold set to 0.5. As shown in Table \ref{tab:object-detection-precision-recall}, \GeoFusion outperforms all the baseline methods under different mAP metrics. With the ratio of IOU increases for checking true positives, the performance gap between \GeoFusion and baseline methods becomes larger. Interesting, the recall of the Frame-by-frame method under $\text{mAP}_{50}$is higher than \textit{GeoFusion}. This is because \GeoFusion has a strict way to filter out false and noisy detections. In other words, detections are chosen very conservatively by \GeoFusion which leads to more false negatives in a looser metric. 
\end{paragraph}

\subsubsection{Pose Accuracy} For 6D object pose accuracy, we use the ADD-S metric \cite{xiang2017posecnn} to compute the average point distance between estimated pose and the correct pose. Only true positives are computed at each frame for each method. As our aim is to build a high accuracy object-level map and the precision required by the robot to grasp objects are relatively high, we plot the accuracy-threshold curves within the range of [0.00m, 0.02m] as shown in Fig. \ref{fig:pose_acc}. \GeoFusion achieves the highest pose accuracy among all the methods and about 90\% percent of the objects have achieved pose error under 1cm. The pose accuracy of baseline SLAM method is even worse than the Frame-by-Frame method indicating that the confidence scores from perception module are sometimes not reliable. 



\subsubsection{Run Time}

\begin{figure}[t]
    \centering
        \includegraphics[width=0.9\columnwidth]{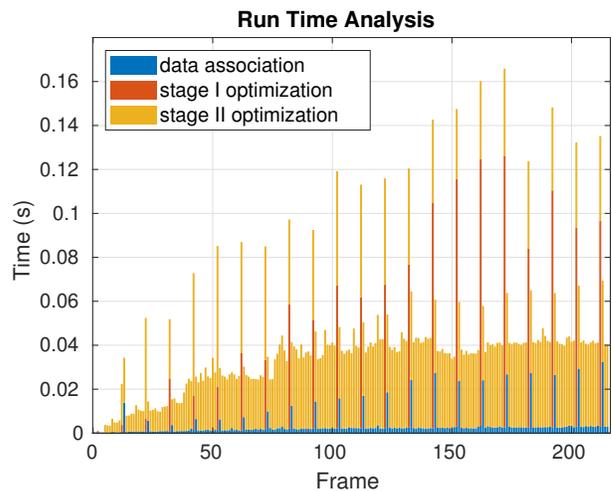}
    \caption{\footnotesize Time plot for each frame of the scene in the Fig. \ref{fig:teaser}} 
    \label{fig:time_plot}
\end{figure}

We show an analysis of run time in Fig. \ref{fig:time_plot} on each frame for data association and optimization of an example scene in Fig. \ref{fig:teaser}. Note that our method only runs on CPU (without measurements generation), the average time for each frame is under 200 ms


\subsubsection{Geometric Relation} 
\begin{figure*}[]
    \centering
        \includegraphics[width=0.45\columnwidth]{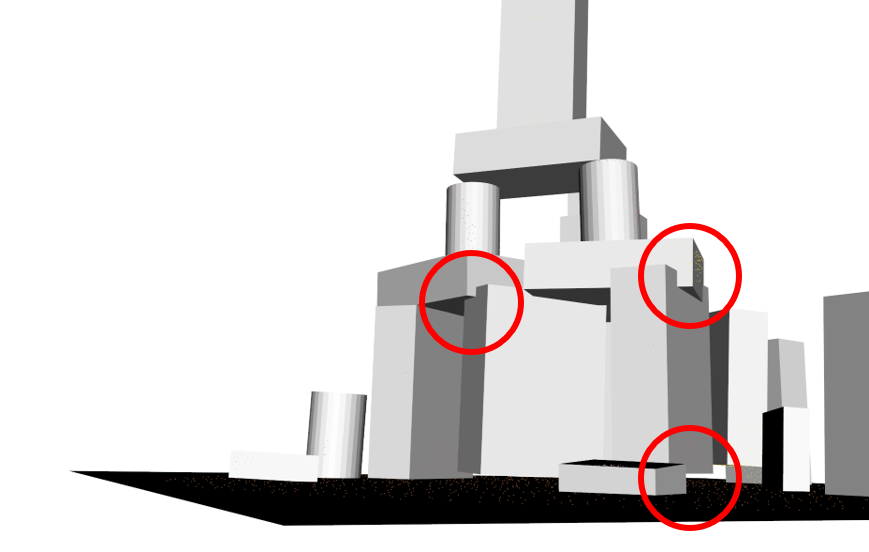}
        \includegraphics[width=0.45\columnwidth]{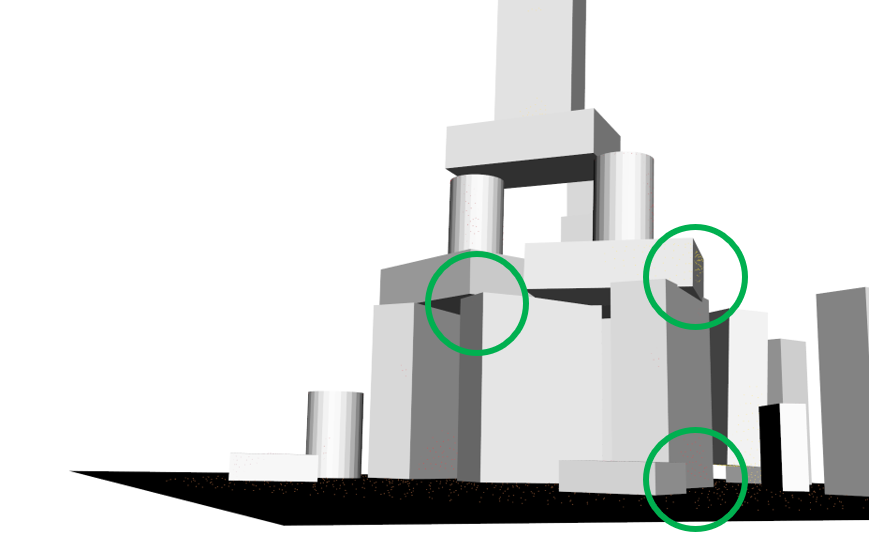}
    \caption{\footnotesize The comparison of the two-stage optimization of 230th keyframe of the scene in Fig. \ref{fig:teaser}.
    } 
    \label{fig:geometry_relation_compare}
\end{figure*}

Fig. \ref{fig:geometry_relation_compare} compares the results of two optimization stage of the scene in Fig. \ref{fig:teaser}. Left figure is the result of optimization stage I without geometric relation constraints and the right figure is the result after optimized over geometric constraints. The geometric constraints pulled objects to satisfy the constraints. For example, the objects on the right are more aligned with the table and the intersection between objects is fixed. 

\section{Conclusion}
In this work, we present \textit{GeoFusion}, a SLAM-based scene understanding method for building a semantic map at object-level in dense clutter while taking into account geometric consistency. Our method is able to infer object labels and 6D poses from noisy semantic measurements robustly and efficiently. The reasoning at object-level with geometry offers a fast and reliable way to filter out false positives and constraint the object through geometric relation. The computed geometric relations are also directly amenable to high-level task planners for robots to reason over actions for goal-directed manipulation tasks. 







\bibliographystyle{abbrv}
\bibliography{big}





\end{document}